# Personalized and Aspiration oriented career path finder by modelling similarity of career paths

Kuleshwar Sahu[1], Girish Keshav Palshikar[1]and Rajiv Srivastava[1,2]

[1] Tata Research Development and Design Center, TCS, 54-B, Hadapsar Industrial Estate, Hadapsar, Pune, INDIA
[2] Department of Technology, University of Pune, Pune, INDIA
{kuleshwar.sahu, gk.palshikar, rajiv.srivastava}@tcs.com

**Abstract.** Fulfilling career aspirations is important for growth of employee and x`organization. We propose a data driven methodology to recommend personalized career path for a given aspirant's career path and aspirations. The proposed method uses the career path similarity (CPS) between aspirant's career and candidate career path, and 'aspirational similarity' (AS) between aspiration and candidate career paths to find suitable career path. CPS ensures personalized recommendation while AS ensures aspiration fulfillment. We defined two methods to compute the CPS between career paths which are (a) domain knowledge driven (DKD) and, (b) unsupervised representation learning and alignment (URLA) based, along with different AS measures. The DKD based similarity is defined in the terms of features extracted and summarized over career paths. In the URLA, we use the sequence of event names present in the career paths of the employees to learn the embedding for each event name. In URLA we use learned embedding vector of the career path event names and associated event attributes (skill cluster and domain) to find the best alignment between two career paths. We hypothesized that relative position of event names in the sequence represents semantics of event name and that can be learned. We use LSTM neural network to learn the embedding vector of each career event name. We also define the matching method to compute the AS between aspiration and career path in both proposed methods. We combine CPS and AS to rank available 'candidate career paths' of employees to find the suitable one. We get better DCG value in URLA as compare to DKD. We also showed that ranking are coherent using both the methods. URLA method is better since it does not require domain knowledge to model the similarity and includes temporal aspect by optimal Levenshtein alignment using weighted cosine distance. As per our knowledge, it is first ever unsupervised method to recommend a career path by taking explicit aspiration into consideration.

## 1 Introduction

Fulfilling career aspirations is important for growth of employee and organization. In the information technology (IT) domain, the landscape of technologies, associates tasks, roles and service offerings are continuously expanding. Given this pace of change and proliferation of technology areas and roles, there is a significant increase in the available career options. The organizations are actively training and re-training employees in new technologies [Learning recomm. paper], associated tasks, and application areas to plan and fulfill the upcoming market demands.

A satisfied employee leads to a satisfied customer, and for an employee to be satisfied with her career, organization should help her plan, follow and meet her aspirations. A career path is a sequence of events such as project allocation or role change over the entire tenure in the firm. eHRMs, the workflow automation systems, gather rich data about career events of all the employees. The organizational Learning and Development (L&D) groups are encouraging voluntary disclosure of employee aspirations. The aspiration driven career advice, if personalized using past career profile would have higher acceptance of employee. The timely advice would lead to engaged and satisfied employees, reducing voluntary turnover. The generated career advisory can also guide employee allocations [optimal allocation paper]. The Human Resource Management (HRM) departments of large services organizations are deploying career advisory systems to guide and mentor employees. The career guidance includes suggesting multiple work related events having multiple attributes such as technology, role, domain, industry, learning as well as employee aspirations such as location preference, financial targets, along a timeline.

The career profiling and providing advisory is a challenging problem, it can benefit from the eHRM data. A popular approach for generating career advisory for a candidate (employee) is to identify 'similar' but 'senior' employees who have attained the aspirations specified by the candidate. Another important application, which is similar in character, is for finding 'similar' candidate profiles or resumes [Candidate matching paper] for a job, given an 'ideal' employee profile. Both these applications treat career as a sequence of events such as project allocation or a job assignment, transition of roles, acquisition of skills, trainings or certifications, etc.

We propose two data driven methods to recommend personalized career path for an employee in an industrial setting, given her career path and aspirations. Given an organization repository of employee profile data, consisting of multiple project assignments, skills, role, domain and learnings, the proposed method generates the multi-attribute event based personalized recommendation. This algorithm is novel as it caters to the specified aspiration of the employee. Both the proposed methods are unsupervised in nature. The first method uses the domain knowledge to evaluate the 'similarity' of the career profiles, whereas the second method is domain independent which learns the representative embedding vectors for the events present in career path.

### 1.1 Career path and similarity notions

**Career path**

On joining the industry, employee's career path starts. During the tenure at industry employee gets essential training and certification to acquire some skills. Upon getting required skills she is allocated to a project to perform certain tasks defined by the role and responsibilities, she is assigned to particular domain of application. For instance, an employee joins IT services company "XYZ", gets initial training and certification related to "Java" technology and is allocated to a project "PQR" in "Banking and Finance" domain, in 'Developer' role to build application for banking customers. We define a career path of an employee using four type of events in chronological order as depicted in the Table 1 and 2. A career path of an employee consists of four essential type of events as 1. Project Allocation. 2. Role Assignment. 3. Training and 4. Certification.

**Table 1.** Sample career path

| Project Allocation | 'Project 1" | 'Project2' | 'Project1' |
|---|---|---|---|
| Role Set | Developer | Developer | Module Leader |
| Training | \| 'Java Foundation' | \|'J2EE- advanced' | \|'Hibernate' |
| Certification | \| 'Sun Java Certification' | | \|Stanford Lead |
| Calendar time | \|01 Apr 17 \|01 Oct 17 \|01 Apr 18 \|01 Oct 17 \|01 Apr 19 \|01 Oct 17 | | |

**Table 2.** Sample career path along with event attributes

| Project Allocation | {'java', 'oracle'} ('Banking and Finance') | {'java', 'accounting'} ('Banking and Finance') | {'java', 'oracle'} ('Banking and Finance') |
|---|---|---|---|
| Role Set | Developer | Developer | Module Leader |
| Training | \| {'java'} (IT) | \|{'java'} (IT) | \|{'java'} (IT) |
| Certification | \| {'java'} (IT) | | \|{'Soft Skill'} (HR) |
| Calendar time | \|01 Apr 17 \|01 Oct 17 \|01 Apr 18 \|01 Oct 17 \|01 Apr 19 \|01 Oct 17 | | |

**Career Path Similarity**

A career path is a sequence of events and corresponding career events over the entire tenure in the firm. The similarity between two career paths will be defined using career path events and event attributes. We define the career path similarity using four events and their attributes as *project_similarity*, *role_similarity*, *domain_similarity* and *learning_similarity*. We combine these four similarity measures to compute the overall career path similarity measure. We compute each of these four similarity components using the two proposed methods:

1. Domain knowledge driven (DKD): In this approach using the domain knowledge of the career path progression, we extract important cumulative features of the career path using which we define similarity measures.

**2.** Unsupervised, representation learning and alignment (URLA): We compute the sequence-to-sequence similarity over learned embeddings of the event names using weighted edit distance.

In DKD method, most of the similarity computations are based on the exact matching of events or associated attributes such as role ‘Module Leader’ does not match with ‘Project Manager’ and similar restrictions are in matching of other events or associated attributes. As we know that, there is an overlap between task profiles of ‘Module Leader’ and ‘Project Manager’, the implementation of a comprehensive function to evaluate partial, granular, and non-discrete similarity is very difficult for any domain. In the URLA approach, we overcome this limitation by learning the embedding for the each event or event attribute which is used and compute similarity using uniform method of cosine similarity.

## 2 Related work

There is work done in cumulative profile as well as sequential profile similarity for various applications. Earlier work reported in the recruitment context [5] as only ‘next job prediction’ problem solved using a supervised approach, considering candidate profile, organizational profiles and job transition data. Another approach in [4] addresses ‘similar’ profile match, using logistic regression, considering model of career trajectory consisting of job nodes. Each job node has attributes such as current job title, company, size of the company, current functions, job seniority, etc. The weights of recent job events decay exponentially to accommodate temporal importance. A dissimilar event node is a ‘gap’ that is penalized. An important shortcoming of such approaches is that these completely discard ‘not similar’ events in a sequence. The [11] proposes a discriminative probabilistic model that identifies latent content and graph classes for people with similar profile content and social graph similarity patterns, and learns a specialized similarity model for each latent class in the context of use of social media use for identifying professional similarities.

The current work [2] suggests smooth or non-discrete transitions across job or assignment nodes, which learn embedding vectors using the event data sources from social media. The problem is modeled as multi-source, multi-task framework to achieve temporal smoothness where in a least-square error optimization technique is used to learn the weight vector for attributes consisting of demographics, topics, LIWC features, sourced from the candidate’s social media footprint to predict her next role. These methods do not take into account aspirations of the candidates. For the multi-modal sequence matching for patient record similarity [1] uses, dynamic matching of temporal patterns in patient sequences using representation embeddings learnt using Recurrent Neural Network (RNN) architecture adopting ‘Word2Vec’ technique. The [16] represents the EHRs for every patient as a temporal matrix with time and events as dimensions. They build a four-layer convolutional neural network model for extracting phenotypes and perform risk prediction for patients.

Golbeck in [12] shows feature extraction from the social, cumulative profile attributes which correlate and help in trust levels derivation, to represent similarity, amongst users in social recommendation scenario. The [15] predicts the additional skills for the profile using collaborative filtering to improve the profile for matching for offered services of LinkedIn.

For the recruitment domain, several companies offer products that use information retrieval methods for ranking using the candidate profile, job description as well as candidate's interest in job change [13]. These products also have a feature of search based on ideal candidates. The [14] extends the concept of homophily to construct personalized features from LinkedIn profile content to find similar profiles. For example, the expertise homophily, authors propose an approach incorporating members' profiles, skill-endorsement graphs and skill co-occurrence patterns to estimate their expertise scores for the skills.

Our approaches consider aspiration as an input to recommend a suitable career path. The techniques used are unsupervised, and use representation learning from the data generated using negative sampling. One of our method is domain independent and gives comparable recommendation quality, which can be used for recruitment.

## 3 Problem Formulation

We are given set of career paths $C= \{c_1, c_2, c_3, \ldots.., c_n\}$ of the employees. A career path $c_i \in C$ can be represented by list of time ordered tuples. Each tuple is represented by four elements as (Event type, event name, start date, end date). Event type can be either of {'Project Allocation', 'Role Set', "Training', 'Certification'}. Each tuple represents an event in the career. Each event name in the event associated with event attributes as {skill cluster set, domain} except "Role Set" event type. We identified subset of career paths as $D \subseteq C$ such that each $d_i \in D$ can be split out into two career paths as current career path (must have a current role) and future career path (must have complete aspiration information). We split first 30% by duration as current career path and remaining 70% as aspirational career path. For each career path in $d_i \in D$, we create current career path and aspiration tuple as $(t_i, a_i)$. Consequently, we create a set of tuple of career path and aspiration $D' = \{(t_1, a_1), (t_2, a_2), \ldots, (t_m, a_m)\}$. Each $a_i$ is represented by a tuple of three elements as (Role name, domain, {skill clusters}). For each tuple in $(t_i, a_i) \in D`$, we create candidate list of employee whose career path can be recommended using some rules. These rules are, candidate career path should have attained the aspired role present in $a_i$ and total experience of candidate career path should be more than $\Delta$ (2.5 years) years than the current career path $t_i$ We formed the set $D`` = \{(t_1, a_1, l_1), (t_2, a_2, l_2), \ldots, (t_m, a_m, l_m)\}$, where each $l_i \subset C$ represents list of candidate career paths for current career path $t_i$ , and aspiration $a_i$. The problem is to find a suitable career path $c_i \in l_i$ for $i^{th}$ current career path and aspiration tuple $(t_i, a_i)$. It can be stated as a ranking problem to rank the candidate career paths and report the best career path. Proposed solutions are discussed in the subsequent section.

# 4 Solution approaches

We used similarity as measure to find the suitable career path for a given current career path and aspiration. If aspirant's current career path is more similar to the candidate's past or present career path's characteristic then this candidate career path would be reasonable choice to be recommended. We compute career path similarity between current career path and candidate career path to ensure that recommended career path would be easier for the aspirant. We later used aspiration similarity among the aspiration and candidate career path to ensure that stated aspiration were sufficiently present in the recommended career path. We average career path and aspiration similarity to rank the candidate career paths. We report top ranked career path as best career path for the given aspirant. We proposed two methods to rank the candidate career paths, we used domain knowledge from *Employee Talent Management System* to come up with an algorithm to find the best career path, we call this method as domain knowledge driven (DKD) career advisory. It has got advantage of domain expertise to model practical aspect in the career path recommendation, detailed in the subsequent section. Domain knowledge driven proposed framework is based on the exact matching between the event names hence does not allow any sort of partial matching. Moreover, domain knowledge driven approach does not consider the temporal alignment in the career paths. It does not consider temporal aspect present in the career path. The proposed second method overcomes above stated two limitations by learning the embedding for each event names in the career path and then using alignment based method we find the best candidate career path. We name this method as unsupervised representation learning and alignment (URLA) based career advisory.

## 4.1 Domain knowledge driven career advisory

In this approach we define both career path similarity and aspiration similarity measure using domain knowledge. We define non-symmetric similarity measure from aspirant career path to candidate career path. For similarity measurement apart from events names matching, we also consider attained duration to get more reasonable estimate of the similarity. For instance, ('Developer', '500 days') is more similar to ('Developer', '600 days') than ('Developer', '300' days). We use trapezoid function of attained duration of candidate to aspirant career path as a weighing coefficient over event names matching score. Using trapezoid function in above example algorithm will select ('Developer', '600 days') as better match.

*Career path similarity* between aspirant career path and candidate career path, consists of following distinct four similarity components:

1. *Learning similarity:* It is intersection of the courses and certification present in aspirant's current career path to candidate career path, it is represented by $Sim_{CourseCert}$ in *Figure* 1 of proposed algorithm.
2. *Role similarity:* It measures the similarity in terms of common roles along with durations of aspirant and candidate career path and denoted as $Sim_{Role}$.

3. *Project similarity:* It is measures the similarity in terms of the present common skill clusters in projects, number of projects and duration of the projects. Project similarity is denoted by $Sim_{Project}$.
4. *Domain similarity:* It measures the similarity by number of projects in similar domain and duration attained by aspirant to candidate career path and denoted by $Sim_{Domain}$.

*Aspiration Similarity* is computed between aspiration information and candidate career path. Since aspiration information contains three elements viz. role name, domain and skill cluster set, hence aspiration similarity also consists of the following three aspiration similarity components.

1. *Role similarity***:** It measures similarity in terms of presence of aspired role in candidate career path along with duration and also distinct projects worked in aspired role by candidate career path.
2. *Domain similarity*: It measure the similarity by the number of projects from the aspired domain along with duration and also distinct projects worked in aspired domain by candidate career path.
3. *Skill cluster similarity***:** Skill cluster information is present in three event types viz. 'Project Allocation', 'Training' and 'Certification' hence we define three sub-similarity component which are:
   - *Project skill clusters similarity***:** Stated skill clusters in aspiration are matched with project event's skill clusters along with duration to measure aspired skill clusters presence in the project career path of the candidate.
   - *Training skill cluster similarity*: It measures proportion of the skill clusters mentioned in the aspired skill cluster which are present in the trainings presents in the candidate career path.
   - *Certification skill cluster aspiration similarity***:** Similar to above it measure similarity for certifications done by the employees.

Fig 1 denotes the similarity functions of the domain knowledge driven career advisory method. The method *career_state_to_features* returns the tuple (*dL, rL, cL, sL*) which is (list of domains, list of roles, list of courses and certifications, list of skill clusters) from the career path. The algorithm takes input as aspirant's current career profile and list of candidate career paths. Method *career_path_sim* computes the career path to career path similarity as discussed. After that algorithm removes some irrelevant event present in the candidate career path. For instance, if candidate career path contains the aspirant's current role then events before that are obsolete since aspirant has already reached that role. We perform method *Aspirational_Similarity* to get the aspirational similarity between aspiration and refined career path. We get career path and aspirational similarity for all candidate career paths with aspirant's current career path and aspiration, both similarities are normalized in range [0,1]. We took average of both similarity measures to rank the candidate career paths and report top ranked career path as best career path.

### 4.2 Unsupervised-Representation learning and alignment based career advisory

Above domain knowledge driven career advisory method has certain disadvantage as it does not compute any similarity among different event value/names i.e. does not allow domain specific semantic matching. Due to this restriction, above approach does not search for the other suitable career path which are partially matching career path, but may be better in real life scenario since its impractical to find an exact matching career path. For instance, it assume zero similarity among 'Developer' and 'Module Leader', similarly it assumes zero similarity between the skill cluster 'java' and 'oracle', and it also assumes 'Banking and Finance' completely different from 'Insurance' domain. It also does not consider the temporal aspect present in two career paths. For instance, role sequence {('Developer', 300 days), ('Module Leader', 500 days)} would be better aligned with {('Developer', 325), ('Module Leader', 600)} than {('Developer', 325), ('Tester', 200), ('Tester Analyst', 334), ('Module Leader', 600 days)} but domain knowledge based method will treat both as similar. We incorporate partial and semantic matching using the matching in embedding space. We find best temporal alignment to search most relevant similar career paths.

**Learning Embedding of career path event names/values:** Data set contains sequence of four distinct type of events. We can divide a career path into four event specific sub-career paths of different types as an employee goes through sequence of project allocations, role assignments, trainings and certifications. As per [10], such segregation of event types reduces the effects of data sparsity in multi attribute data.

We incorporate following hypotheses to use embeddings of event names as a proxy for the event names

- First hypothesis is that an employee is able to transition from one project to another project then both projects must possess some similarity.
- Second hypothesis is that project which are temporally nearer in the sequence are more similar to each other as compared to temporally aligned project which are farther. Employee's experience and skill set changes over period of time hence he would be capable enough to handle different types of projects later seen in the sequence i.e. first project and second project would more similar than first and tenth project in the sequence.
- Third hypothesis is that, project which are not directly or mutually present in the same sequence are more different than the project which are from the same sequence. Given three sequences ($p_1$, $p_2$, $p_3$) and ($p_2$, $p_7$, $p_8$) and ($p_{10}$, $p_{11}$) learned embedding will ensure $p_1$ and $p_7$ are more similar than $p_1$ and $p_{10}$ since $p_1$ and $p_7$ are either directly or mutually related.
- Similar hypotheses for other event types.

**Function** $sim_{\text{CourseCert}}(S_e, S)$
$(dL, rL, cL, sL)$ := *career_state_to_features*$(S_e)$;
$(dL1, rL1, cL1, sL1)$ := *career_state_to_features*$(S)$;
**return** $\frac{|cL \cap CL1|}{|CL \cup CL1|}$

**Function** $sim_{\text{Role}}(S_e, S)$
$(dL, rL, cL, sL)$ := *career_state_to_features*$(S_e)$;
$(dL1, rL1, cL1, sL1)$ := *career_state_to_features*$(S)$;
**return** $\left(\frac{1}{2} \cdot \left(\frac{1}{|rL|} \sum_{r \in rL} trapezoid\left(\frac{\#projects\ in\ role\ r\ in\ S}{\#projects\ in\ role\ S_e}\right) + \frac{1}{|rL|} \sum_{r \in rL} trapezoid\left(\frac{Total\ duration\ of\ projects\ in\ role\ r\ in\ S}{Total\ duration\ of\ projets\ in\ role\ r\ in\ S_e}\right)\right)\right)$

**Function** $sim_{\text{Project}}(S_e, S)$
$(dL, rL, cL, sL)$ := *career_state_to_features*$(S_e)$;
$(dL1, rL1, cL1, sL1)$ := *career_state_to_features*$(S)$;
**return** $\left(\frac{1}{|sL|} \cdot \frac{Total\ duration\ of\ projects\ in\ S\ which\ use\ at\ least\ one\ skill\ in\ sL}{Total\ duration\ of\ projects\ in\ S} \cdot \sum_{s \epsilon sL \cap sL1} trapezoid\left(\frac{Total\ duration\ of\ projects\ in\ S_e\ in\ skill\ s}{Total\ duration\ of\ projects\ in\ S\ in\ skill\ s}\right)\right)$

**Function** *career _path_sim*$(S_e, S)$ // this is NOT a symmetric function – order of arguments is important
$(dL, rL, cL, sL)$ := *career_state_to_features*$(S_e)$;
$(dL1, rL1, cL1, sL1)$ := *career_state_to_features*$(S)$;
s := $sim_{\text{CourseCert}}(S_e, S) + sim_{\text{Domain}}(S_e, S) + sim_{\text{Role}}(S_e, S) + sim_{\text{Project}}(S_e, S)$;
*s* := s / 4; // bring *s* to a number between 0 and 1
**return** *s*

*F*: refined career pat; *r* = aspired role; *d* = aspired domain; *slist* = list of skills in specified skill clusters
**Function** *Aspirational_Similarity*(*F*, *r*, *d*, *slist*)

$$s := \frac{\#courses\ in\ domain\ d\ in\ F}{\#MaxDomainCourses} + \frac{\#certifications\ in\ domain\ d\ in\ F}{\#MaxDomainCertifications} + \frac{\#projects\ in\ domain\ d\ in\ F}{\#projects\ in\ F} + \frac{Total\ duration\ of\ projects\ in\ domain\ d\ in\ F}{Total\ duration\ of\ projets\ in\ F} + \frac{Total\ duration\ in\ role\ r\ in\ F}{Total\ duration\ of\ F} + \frac{\#projects\ in\ role\ r\ in\ F}{\#projects\ in\ F} + \frac{Total\ duration\ of\ projects\ in\ role\ r\ in\ F}{Total\ duration\ of\ projets\ in\ F} + \frac{1}{|slist| \cdot Total\ duration\ of\ all\ projects\ in\ F} \sum_{i=1}^{\#projects\ in\ F} (\#skills\ from\ slist\ used\ in\ i-th\ project\ in\ F) \cdot Duration\ of\ i-th\ project\ in\ F)$$

**return** s/8

**Fig. 1.** Similarity functions used in the DKD algorithm

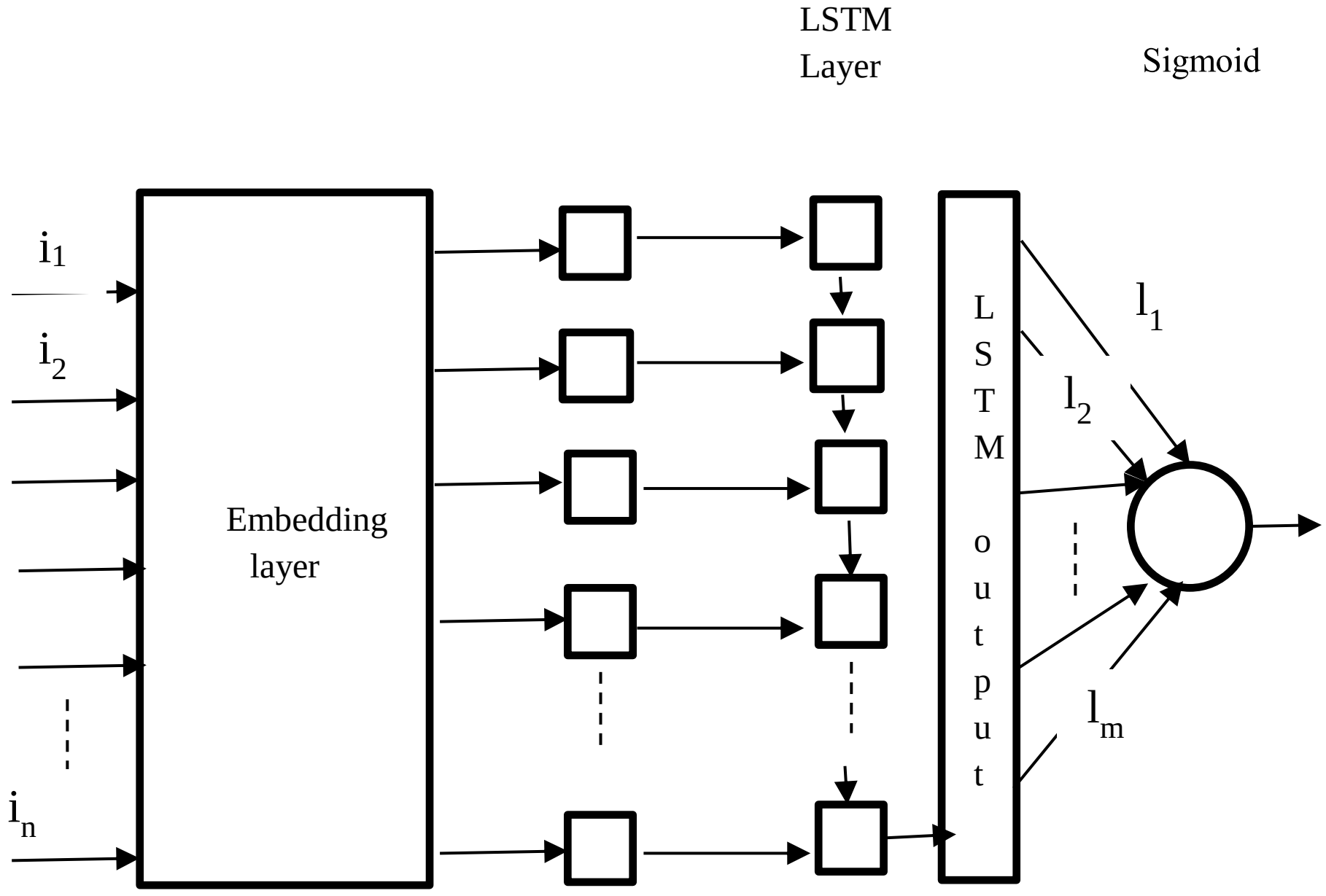

**Fig. 2.** LSTM based neural network used to learn the embedding of event names.

*Learning the embedding of the projects:* We have 24184 career paths and same number of project sequences. On average each career path contains 4.37 project allocations. All career path contains distinct 18876 projects (vocabulary size) for which we learn the embedding. We used negative sampling technique [17] to learn embedding vector for each project name using the available valid project name sequences. We randomly generate 24184 invalid project sequence from the vocabulary and have equal number of valid project sequences. We built LSTM network in classification setting for valid and invalid sequence as presented in the Fig **2**. The used deep neural network contains an embedding layer, a LSTM layer, and last output layer with sigmoid function as presented in the figure Fig 2. We trained the model with cross fold accuracy of 87% for the valid and invalid sequence classification. We used the learned embedding for each project names from the embedding layer. The deep neural network in Fig 2 learns the embedding for $n$ (18876) distinct project names in $m$ (30) dimensional vector space using the 24184 valid and invalid project sequences.

Using the similar method we learned the embedding for the domain names, roles and skill clusters with training accuracies of 98.62, 99.98 and 99.98 respectively. Since we have only 19 distinct domain names hence the length of domain sequences are different than project sequences. We removed repeating domain names i.e. ('Insurance', 'Insurance', 'Banking and Finance', 'Banking and Finance', 'IT') which would be converted to ('Insurance', 'Banking and Finance', 'IT'). There are 300 distinct skill clusters. Each project are mapped to more than one skill clusters. We created skill cluster sequence by putting the skill clusters in place of project in project sequence. We have distinct 358 roles for which we have learned embeddings.

*Learning embeddings of the training names:* We learned the embeddings for the training names in two phases. In *phase* 1 we learn the embedding for the training name using words in the training. We supply *phase* 1's learned embedding vectors as initial embedding vectors to *phase* 2 deep neural network. In *phase* 2 learning we use sequence of trainings done by the 24184 employees to refine the embedding.

- *Phase 1 embedding vector using words in training names:* We have got about 3287 distinct trainings available in the learning platform. Each training name is in the form of sequence of words (after removing stop words) such as "Java Foundation course beginners". These all training names contains total 3792 distinct words. Using negative sampling technique [17] we can form another 3287 invalid training names. We used similar LSTM network to learn the embedding of word in the training name with cross fold accuracy of 87%. We took element wise average of learned vectors of each word present in the training name to form a single vector for a training name. We got embedding vector corresponding to each training names.
- *Phase 2 embedding vector using sequence of trainings done in career paths:* After *phase 1* we got embedding vector for each training name. We have the temporal sequence of trainings done in career paths. We can refine embeddings of training names using the sequence information present in the training sequences done by the employees. We hypothesize that trainings which are in temporal proximity must possess similarity. We got around 7198 (since not all employee go through trainings) valid training sequences and we constructed similar number of invalid sequences using random negative sampling. We built similar LSTM network using classification to learn embeddings for each training name.

We used similar method to learn the embeddings for the certification names.

**Sequence alignment based career path finder:** In the previous section we have learned the embedding vector for all of the event name of all types in the career path. An embedded career path is list of temporally ordered embedding vectors of event names along with duration of the events. Each embedded career path has four sub-paths as "Project Allocation", 'Role Set', 'Training' and 'Certification'. Similar to 'domain knowledge driven' approach, we define career path and aspirational similarity, presented in Fig 3, and as discussed in the following paragraph.

*Career path similarity (CPS):* Career path similarity is computed for each event type specific sub-career paths separately, presented as function *computeCareePathSim* in *Figure* 3. We used Normalized Levenshtein similarity to find the best alignment between two embedding vectors of sub career paths and then combines the similarities of each sub career paths. The basic unit of comparison in proposed work is cosine distance between two embedding vectors and is weighed by the trapezoid of duration from candidate event duration to aspirant event duration. We get minimum distance between two sub-career paths using Levenshtein distance as presented in lower right section of the **Fig** 3, we normalized it in [0, 1] by maximum distance possible between two sub-career paths. Consequently, we get similarity between two sub-career paths by subtracting computed distance from 1 which maximum possible distance value. We compute similarity for all career path components and took average of all four components as final career path similarity.

```
Se = Aspirant e's career path; S: list of candidate
career paths; A_e = Aspiration of employee e
Function Rank_Career_Paths (S_e, S, A_e):
    ranked_list =[]
    for each S_i in S
            cps=computeCareePathSim( S_e, S_i)
            as= computeAspriationalSim (A_e, S_i)
        avg_sim = (cps+as)/2
        ranked_list.append(i, avg_sim)
    ranked_list = SortDescending(Result_list)
    Return ranked_list [0]
```

```
Se = Aspirant e's career path; S_i: a candidate career path
Function computeCareePathSim (S_e, S_i):
    Sub_paths_e= groupByEventType(S_e)
    Sub_paths_i= groupByEventType(S_i)
    sim=0
    For each pair (p, q) in (Sub_paths_e, Sub_paths_i)
        dist = NormalizedLaveishtineSim(p,q)
        sim= sim+ (1-dist)
    return sim/4
```

```
A_e = Aspiration of employee e S_i: a candidate career
path
Function computeAspriationalSim (A_e, S_i):
    a.role= A_e.role
    a.domain= A_e. domain
    a.skill_clusters=A_e.Skill_clusters
    role_sim= computAspRoleSim( a.role, S_i.role)
    domain_sim=computeAspDomainSim(a.domain,
    S_i. project)
    skill_clusters_sim=computeAspScSim(a.skill_clust
    ers, S_i.project, S_i. trainings, S_i. certifications )
    fin_sim= (role_sim + domain_sim +
    skill_clusters_sim)/3
    return fin_sim
```

Cost= trapezoid( $d^j_c$ / $d^i_e$) * cosine_distance ($S^i_e$, $S^j_c$)
dist [$i$][$j$]= min { dist [$i$-1][$j$], dist [$i$] [$j$-1],
dist [$i$-1][$j$-1] + cost }
where $d^j_c$ = duration of $j^{th}$ event of $c$'s (candidate) career path.
$d^i_e$ = duration of $i^{th}$ event of $e$'s (aspirant) career path.
$S^i_e$= embedding vector of $i^{th}$ event of $e$'s (aspirant) career path.
$S^j_c$= embedding vector of $j^{th}$ event of $c$'s (candidate) career path.

**Fig. 3.** URLA based similarity functions

**Aspirational similarity**: Aspiration information contains role name, domain and set of skill clusters. We have already learned embedding for each of these aspiration components. We compute aspiration similarity as similarity between embedding of aspirations to embedding of candidate career path using function computeAspriationalSim. Aspirational similarity is average of three similarity components explained.

*Role related aspirational similarity (computAspRoleSim):* We compute similarity between aspired role embedding vector with each role embedding vector present in the role career path, weighted by trapezoid of ratio of current duration of role to average duration of roles in the sequence. We add all these similarities and divide it by role career path length which is number of events.

*Domain related aspirational similarity (computeAspDomainSim):* Similar to above, we compute similarity between aspired domain embedding to domain sequence embedding.

*Skill clusters related aspirational similarity (computeAspScSim):* Similar to above, we compute aspirational similarity among an aspired skill cluster and skill cluster sequence in the career path and we average of all aspired skill clusters similarities.

We compute average of career path similarity and aspirational similarity to rank the candidate career paths and report the best career path.

## 5 Case study data set

We have 24184 career paths of employees from large IT industry. The data set contains total 250470 events of either type 'Project Allocation', 'Role Set', 'Training', and 'Certification'. Table 3 shows the basic distribution of events of each type. We split each career path into two career paths as first 30% of the events by duration as current career path and remaining 70% of the career path as future career path. We consider current career path as path of the employee who is looking to fulfil aspiration, as present in the future or 70% part of the sequence. We extract the aspiration information from the future career path.

**Table 3.** Career path data summary

| Event type → | Project Allocation | RoleSet | Trainings | Certifications |
|---|---|---|---|---|
| Total number of events | 108869 | 38796 | 100474 | 2331 |
| Average number of events per career path | 5.50 | 1.60 | 4.15 | 0.09 |
| Total allocated in person days | 14936539 | 5578974 | NA | NA |
| Average allocation in person days | 617 | 230 | NA | NA |

We extract aspired domain as dominant domain worked in the future career path. We also extract the current role in future career path as aspired role and set of skill clusters present in the projects work done by the employee as aspired skill clusters. We found 2461 current career paths and corresponding future paths where all aspiration components such as role name, domain and skill clusters are present. For each found current career path and aspiration combination, we search for candidate career path in all available career paths who have worked on aspired role and have experience more than 2.5 years than current career path. We found 2091 current career paths and corresponding aspiration for which we are able to find at least one or more candidate career path other than aspirant career path. Table 4 represents the summary statistics of the number candidate career paths per aspirant employee.

**Table 4.** Candidate career paths distribution

| min | Mean | Q1 | Q2 | Q3 | max |
|---|---|---|---|---|---|
| 2 | 58.06 | 4 | 18 | 66 | 1311 |

# 6 Experimental results

We evaluate the results on large IT employees data set using both the techniques discussed in solution approaches. It is to be noted that the domain knowledge driven (DKD) method reports highest precision@$k$ because it is based on the exact matching, and original profile is very likely to be found compared to URLA as shown in the Table 5. It would of interest to know how good both algorithms find suitable career path other than the original career path. If we take difference between the known original career path similarities to best candidate career path (other than original by similarity) as an error then DKD report mean error 0.117 while URLA reports mean error of 0.017. This concludes that DKD is better in finding the original career path while URLA is better in finding the best candidate with similar and unseen career path. In other words, DKD similarity drops after finding the original career path while URLA retains better similarity in better career paths other than original career path. T-test shows t-value = -33.48 between URLA error to DKD error which says error are lesser in URLA. We also compared the ranked recommendation quality between DKD and URLA method using Discounted Cumulative Gain (DCG) for various ranks as presented in the Fig 4, the mean DCG values are better fort all ranks in URLA as compared to DKD. For use in DCG computation the relevance score of the recommended candidate career path as current candidate career path similarity score / top ranked career path similarity score.

$$relevance(i) = \{ similarity(i) \,/\, similarity(top_ranker) \}$$

We make relevance score 1 for the top ranked career path for all the aspirants. We then compute the DCG value for recommendation made for the each aspirant using both DKD and URLA for top $k$ ranks (1 to 50). We compute average of DCG value for all the aspirants. We could see in the figure 4 (left) that URLA's mean DCG values are higher than DKD's DCG which says recommendations made by the URLA are better or comparable to DKD recommendation. We compared Kendall's tau coefficient between ranks of the DKD and URLA, at rank 5, mean value of the Kendall's tau coefficient is 0.36 i.e. positive value indicates that rankings have significant pairwise concordance.

**Table 5.** Precision@k results using DKD and URLA

| Precision@ | DKD | URLA |
|---|---|---|
| 1 | 0.80 | 0.58 |
| 2 | 0.85 | 0.71 |
| 3 | 0.87 | 0.72 |
| 4 | 0.89 | 0.75 |
| 5 | 0.90 | 0.76 |
| 6 | 0.90 | 0.79 |
| 7 | 0.91 | 0.79 |
| 8 | 0.91 | 0.80 |
| 9 | 0.92 | 0.81 |
| 10 | 0.92 | 0.82 |

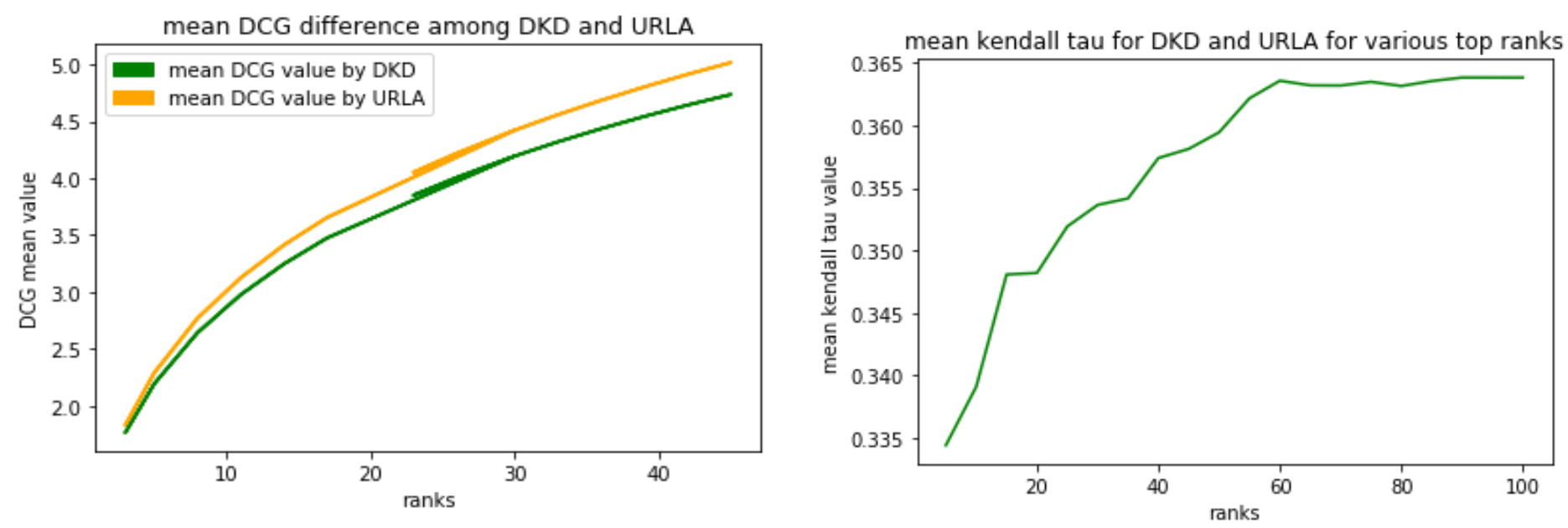

Figure 4 Mean DCG pattern (left) and mean Kendall's tau (right) using DKD and URLA

Figure 4 (right) shows mean Kendall's tau coefficient values at various top ranks (1 to 100), it implies that mean Kendall's tau coefficient is positive i.e. more than 0.33 at any top rank which indicates possible coherence between rankings of the both methods. Observing the above results, we conclude that domain knowledge driven method is better in finding the original ground truth using domain knowledge, but URLA method where we do not require domain knowledge to model the similarity and retains consistent recommendation quality.

## 7 Conclusion and future work

The proposed algorithms recommend a personalized career path having rich set of events along a well depth timeline line. Both the proposed algorithms have good recommendation quality. We start solving the problem by using domain knowledge later we formulated a solution which does not require much domain knowledge which also ensures consistent recommendation quality. The DKD approach is domain knowledge based exact matching method which is better in recalling seen career path effectively as reported by Precision@1 measure. The dependence on rich and granular domain knowledge for building a very accurate similarity function is overcome in the proposed URLA method. The URLA has better DCG as compared to DKD as it learns representation for each event in higher dimensional space improving 'exploration' of the available profile-space. The proposed representation learning based method can be used for the multiple domains and proves to be effective in industrial environment.

As an extension of proposed work we plan to learn a single embedding vector for entire set of temporally ordered events in the career path i.e. a vector embedding for an entire career path. We are also working on problem formulation where ranking of career paths and embedding of career paths can be done simultaneously in a single deep neural network. We are also evaluating ensembles of the proposed methods for better results for different domains such as IT, BPO, Banking, etc. Also the efficacy of the proposed method is being explored for the recruitment related tasks.